\documentclass[biblatex, english]{lni}
\usepackage{booktabs}
\usepackage[]{blindtext}
\graphicspath{{figures/}}
\usepackage{multirow}
\usepackage{subcaption}
\begin{document}
\title{Training a Computer Vision Model for Commercial Bakeries with Primarily Synthetic Images}
\author[1]{Thomas H. Schmitt}{thomas.schmitt@th-nuernberg.de}{}
\author[1]{Maximilian Bundscherer}{}{}
\author[1]{Tobias Bocklet}{}{}%
\affil[1]{Technische Hochschule Nürnberg Georg Simon Ohm\\Center for Artificial Intelligence\\ Keßlerplatz 12\\90489 Nuremberg\\Germany}
\maketitle
\begin{abstract}
In the food industry, reprocessing returned product is a vital step to increase resource efficiency.
\cite{Semmeldetector} presented an AI application that automates the tracking of returned bread buns.
We extend their work by creating an expanded dataset comprising \(2432\) images and a wider range of baked goods.
To increase model robustness, we use generative models pix2pix and CycleGAN to create synthetic images. 
We train state-of-the-art object detection model YOLOv9 and YOLOv8 on our detection task.
Our overall best-performing model achieved an average precision \(AP_{0.5}\) of \(90.3\%\) on our test set.
\end{abstract}
\section{Introduction}
\label{sec:intro}
In industrial processes, keeping accurate track of inventory is vital for smooth and optimal operation.
However, inventory tracking is often a neglected step, especially in reprocessing returned product.
Reprocessing returned products is an especially valuable step in the food industry, which can often reprocess returned products into animal feed or other products.
Furthermore, keeping track of returned products prevents theft and allows for the optimization of production, which in turn increases resource efficiency and boosts profits.
However, keeping track of returned products is often a time and labor-intensive task.
Small to medium-sized companies in particular, struggle with keeping track of returned products, either due to budget constraints or labor shortages.
\cite{Semmeldetector} presented a computer vision application that allows commercial bakeries to track returned bread buns before they are reprocessed into breadcrumbs.
This study expands the scope of their work by:
\begin{enumerate}
\item Expanding the dataset to more images and a wider range of baked goods. 
\item Training the newly released YOLOv9 \cite{YOLOv9} on the baked good detection task.
\item Exploring generative models pix2pix \cite{pix2pix} and CycleGAN \cite{CycleGAN} to generate additional training set images.
\end{enumerate}
\section{Related Works}
\label{sec:related_works}
There are a handful of companies, such as Aiperia \cite{relatedWork_aiperia} and PreciTaste \cite{relatedWork_precitaste}, that offer bakers AI solutions. 
Most, however, focus on business and organizational optimization. 
To the best of our knowledge, aside from \cite{Semmeldetector}, none focus on detecting and differentiating between various baked goods.
\cite{relatedWork_detection_of_bread_3, relatedWork_detection_in_bread_1} employ computer vision models to automate the quality control in the production of baked goods, albeit focused on a specific product and the detection of contamination, respectively.
The problem of training sophisticated detection models on small, highly specialized datasets is most commonly encountered in medical applications.
\cite{relatedWork_medizin_1} trains a CycleGAN model to transform annotated MRI images into CT images to achieve image segmentation on CT images without annotated CT images.
Meanwhile, \cite{relatedWork_medizin_2, relatedWork_medizin_3} employ GAN \cite{GAN} and CycleGAN-based data augmentation to enrich their CT image and X-ray image datasets, respectively.
\section{Data}
\label{sec:data}
\begin{figure}
	\begin{center}
		\includegraphics[trim={0 0.25cm 0 0.25cm},clip,width=0.85\linewidth]{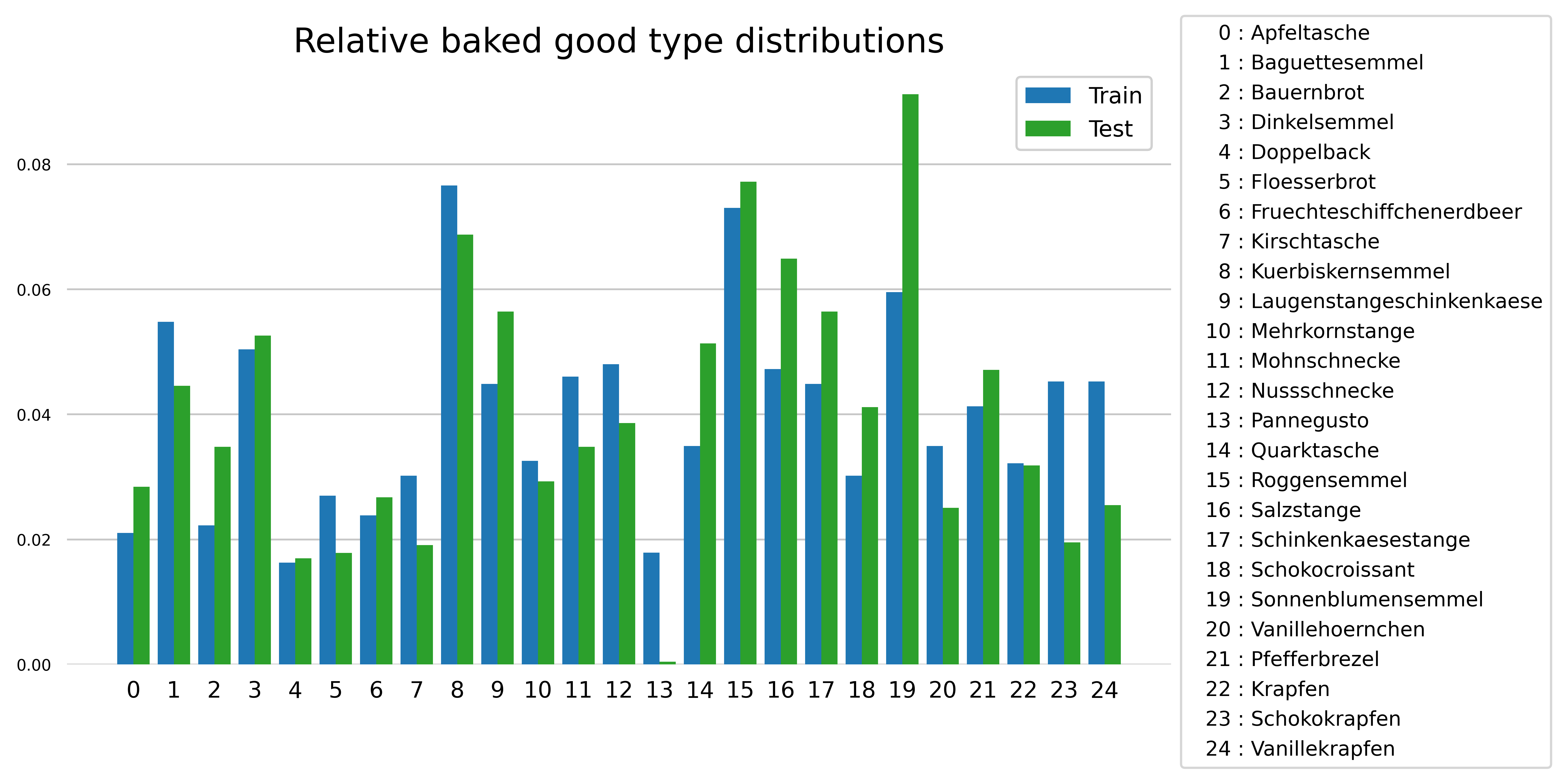}
	\end{center}
	\caption{Relative baked good type distributions in our training, and test set.}
	\label{fig:histo}
\end{figure}

Our dataset comprises \(2432\) images of baked goods divided into \(2050\) training, \(273\) supplementary training, and \(109\) test set images.
Compared to \cite{Semmeldetector} we expand the scope of our datasets to include breads and pastries.
In total, we distinguish between \(25\) different types of baked goods.
The types of baked goods we distinguish in our dataset and the relative baked good type distributions in our training and test sets are shown in Figure \ref{fig:histo}.
\newline
%
Our training set (\(train_{b}\)) comprises \(2050\) images.
Training set images were captured in a relatively controlled environment using an HD webcam.
Each training set image features an individual baked good featuring only a drying tray in the background.
Limiting our training set to images of individual baked goods, facilitates model scalability by speeding up the collection process and allowing us to semi-automatically annotate training set images.
The training set images are captured from a diverse range of camera angles to facilitate model robustness.
This greatly boosts image diversity due to the increased variety of viewpoints and relative object scales.
%
%
Our complementary training set (\(train_{c}\)) comprises \(273\) images.
It comprises images of types of baked goods, which our dataset does not distinguish between.
Each baked good is cast to the complementary baked good type \grqq{}unknown\grqq{} for training.
Our complementary training set serves to increase model robustness by increasing training set image diversity and bolstering resilience against unforeseen baked goods.
To increase robustness against false positives, we use the DIV2K dataset \cite{DIV2K} annotated with empty bounding box annotations as an auxiliary training set (\(train_{a}\)), due to its high image resolution.
\newline
\newline
Our test set comprises \(109\) images of baked goods, with an average of \(21\) baked goods per image.
Test set images were collected by our end user in an alpha test environment using various devices,
this likely makes them truly representative of our use case.
The collection of images directly by our end user in an alpha test environment entails:
(1) That test set images suffer from common image capturing artifacts. (Partial image occlusion, sup optimal image alignment, baked goods in the background)
(2) That a diverse range of baked good samples are featured in our test set images.
(3) That the relative baked good distributions in our test set vary seasonally.
\newline
\newline
Test set images were manually annotated using LabelStudio \cite{LabelStudio}.
Training and supplementary training set images were semi-automatically annotated.
The type of baked good per training set image was manually annotated.
Baked goods were located automatically using the Segment Anything Model \cite{SAM}.
The baked good segmentation masks are given by the biggest non-background segmentation mask per image found by the Segment Anything Model.
A segmentation mask is considered to annotate the background if its corresponding bounding box exceeds an \(IoU\) of \(0.9\) with the entire image.
The resulting segmentation masks are refined with morphology opening and closing operations.
The bounding box annotation is derived from the biggest contour in the segmentation mask.
\section{Image Synthesis}
\label{sec:img_synth}
Following the experience highlighted by \cite{Semmeldetector}, our aim to rapidly scale our models and semi-automatically annotate our training set, significantly diminished the information capacity of our training set.
In preliminary tests, we found, that our training set is insufficient for training large object detection models.
To overcome this limitation, we employ the Copy-Paste augmentation pipeline \cite{CopyPaste} introduced in \cite{Semmeldetector} to create more training set images.
\subsection{Copy-Paste Augmentation}
\label{sec:sec:review}
\begin{figure}
	\centering 
	\begin{subfigure}{0.37\linewidth}
		\includegraphics[width=\linewidth, trim={0cm 0 0cm 0},clip]{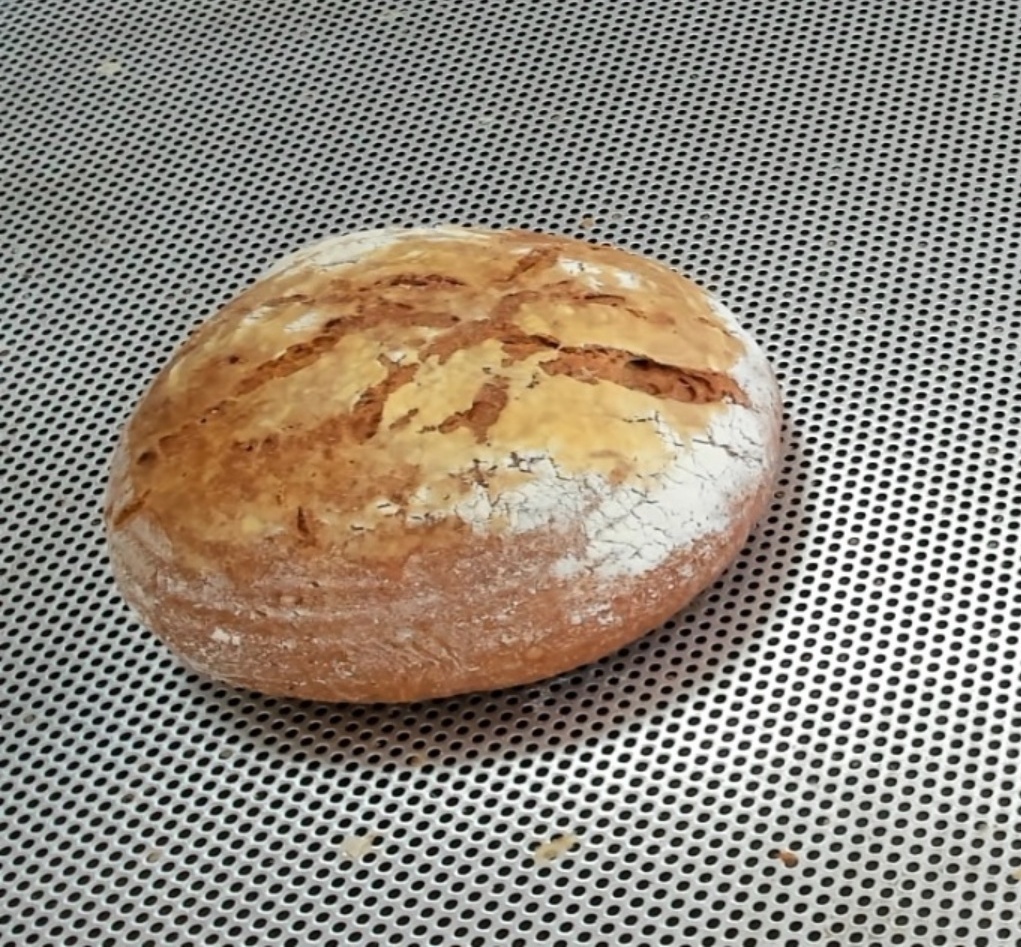}
		\label{fig:examples_1}
	\end{subfigure}\hfil 
	\begin{subfigure}{0.61\linewidth}
		\includegraphics[width=\linewidth, trim={0cm 0 0cm 0},clip]{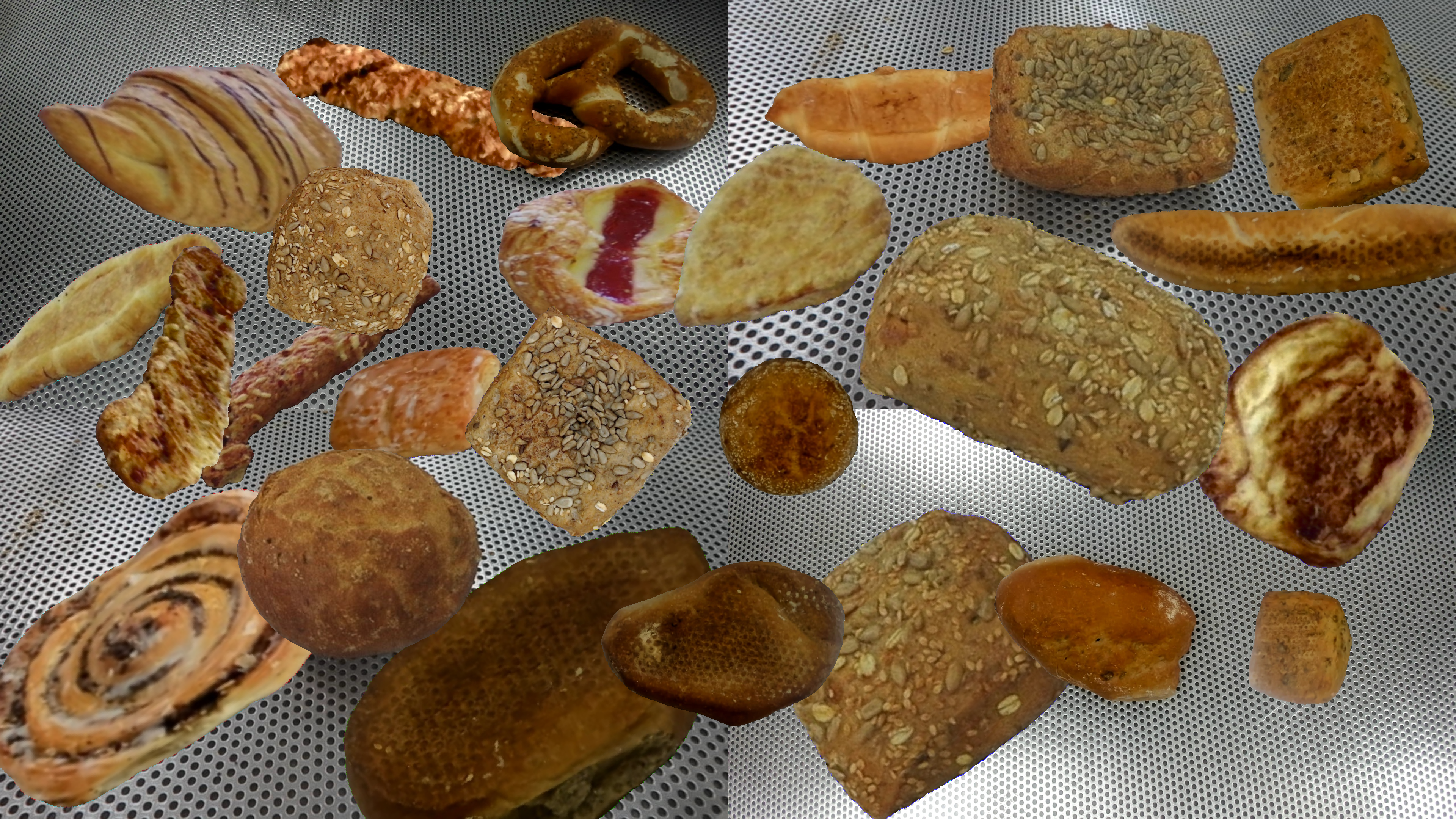}
		\label{fig:examples_2}
	\end{subfigure}
	\caption{Left: Image of a \emph{Bauernbrot} (farmer's bread). Right: Synthetic image generated by the Copy-Paste augmentation pipeline.}
	\label{fig:data_example}
\end{figure}

Using the Copy-Paste \cite{CopyPaste} augmentation pipeline introduced in \cite{Semmeldetector}, we iteratively create crowded images of baked goods, with an average of \(23\) buns per image.
We increase the number of baked goods per image from \(16\) to \(23\) to better reflect our test set images.
To increase the performance of our models on underrepresented baked goods, we balance our synthetic images by oversampling baked goods that accounted for less than \(3\%\) of the training set while synthesizing images.
This affected baked goods: \emph{Apfeltasche}, \emph{Bauernbrot}, \emph{Doppelback}, \emph{Floesserbrot}, \emph{FruechteschiffchenErdbeer}, \emph{Kirschtasche}, \emph{PanneGusto}, \emph{Schokocroissant}.
We limit oversampling to duplicating severely underrepresented baked goods to prevent potential overfitting to particular baked good samples.
Due to the significantly increased range of relative object scales (object size relative to image size) in our training set, we add an object scale check to our image synthesis.
If a baked goods corresponding bounding box accounts for less than \(3\%\) of the synthesized image, it's upscaled to account for \(3\%\); if it accounts for more than \(25\%\), it's down scaled to account for \(25\%\).
This addition helps to control against ill-fitted object scales within one synthetic image.
Baked goods are placed on free spots in the image, determined by the dilated segmentation masks of the baked goods already present in the synthetic image.
To increase image diversity, augmentations: rotation, scaling, low-probability blur, and CLAHE \cite{CLAHE} are applied to the baked goods before they are pasted onto the image.
Background images are generated using a simplified version of the mosaic data augmentation method introduced in \cite{YOLOv4}.
Figure \ref{fig:data_example} shows a image from our training set and one generated by the Copy-Paste augmentation pipeline.
We adopt \cite{Semmeldetector}'s Albumentation \cite{Albu} based online image augmentation pipeline \(DP_{0,04}\) to further improve model robustness,
in which augmentations are applied in a sequential manner to simulate commonly occurring image distortions.
The augmentations are in order: spatial-level transformations (CoarseDropout, PixelDropout, Scale, Rotate), each applied with a likelihood of \(0.01\), followed by pixel-level transformations (Blur, MedianBlur, ToGray, and CLAHE), each applied with a likelihood of \(0.04\).
Bounding boxes are assumed to remain unchanged during augmentation despite the applied CoarseDropout.
\subsection{Generative Models}
\label{sec:sec:gen_models}
\begin{figure}
	\centering 
	\begin{subfigure}{0.32\linewidth}
		\includegraphics[width=\linewidth, trim={0cm 0 0cm 2cm},clip]{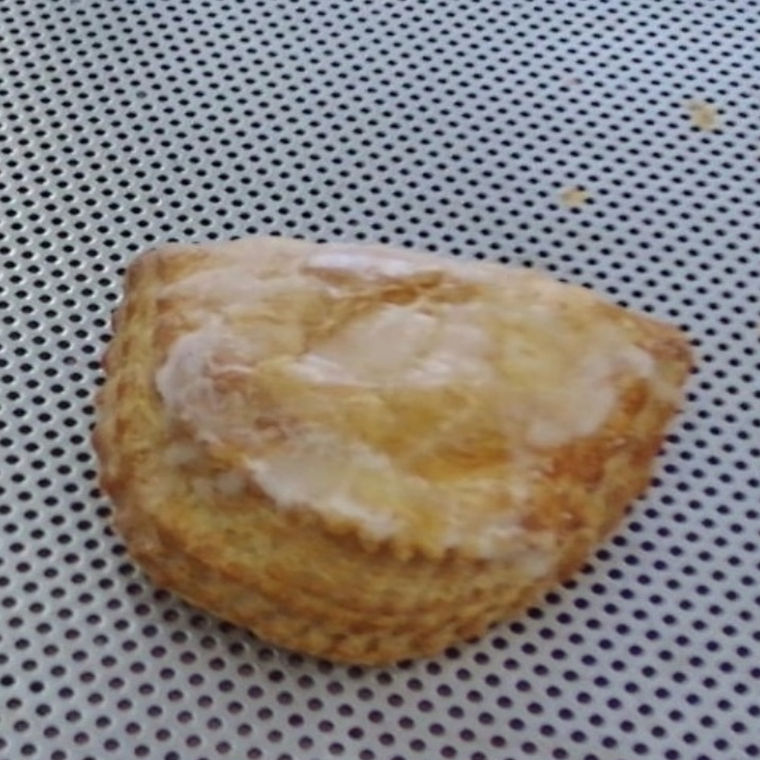}
		\label{fig:examples_1}
	\end{subfigure}\hfil 
	\begin{subfigure}{0.32\linewidth}
		\includegraphics[width=\linewidth, trim={0cm 0 0cm 2cm},clip]{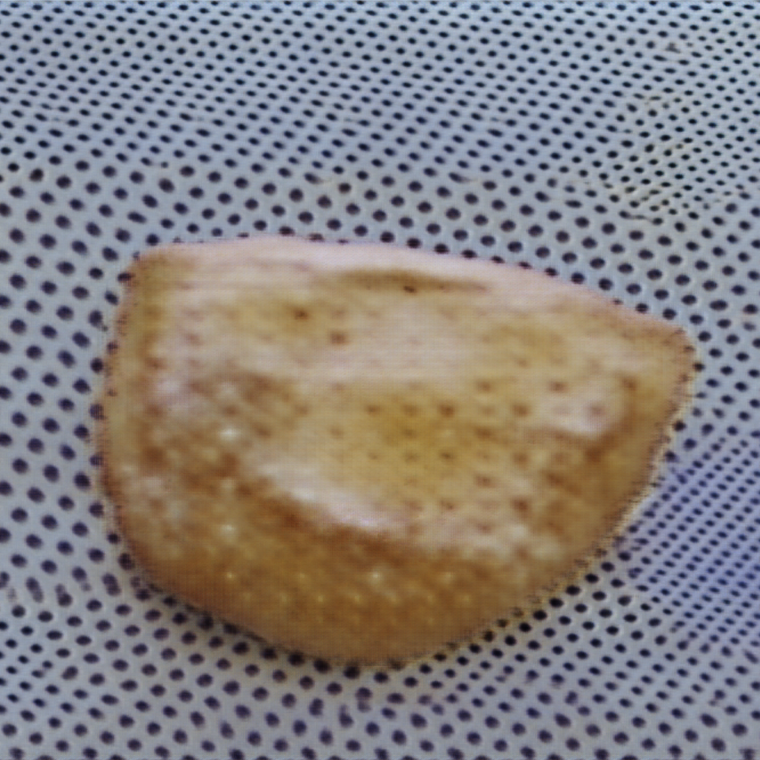}
		\label{fig:examples_2}
	\end{subfigure}\hfil 
	\begin{subfigure}{0.32\linewidth}
		\includegraphics[width=\linewidth, trim={0cm 0 0cm 2cm},clip]{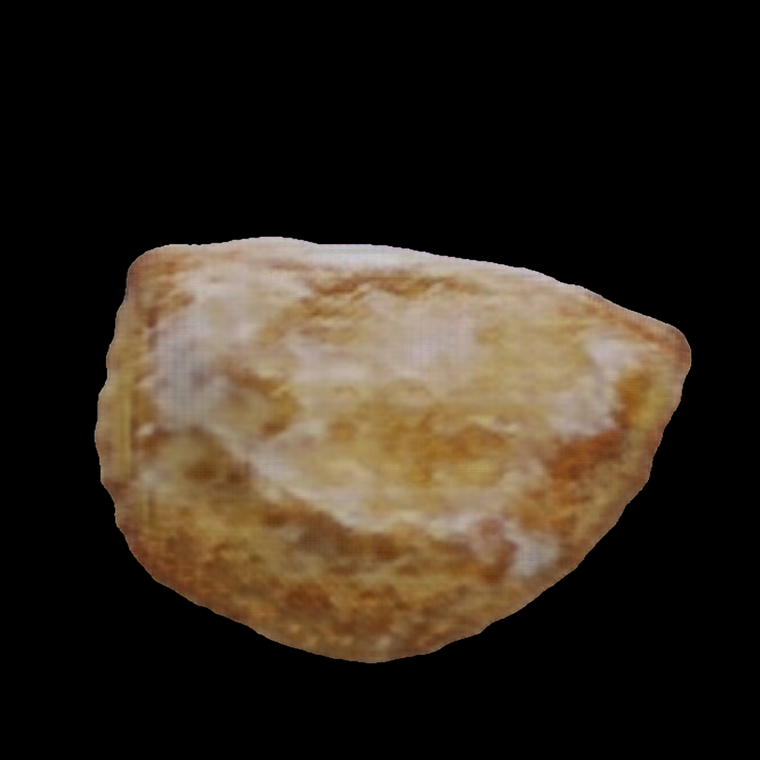}
		\label{fig:examples_3}
	\end{subfigure}
	\caption{Left: Original image of an \emph{Apfeltasche} (apple turnover). Middle: The corresponding synthetic image generated by a model trained on images with a drying tray background. Right: The synthetic image generated by a model trained on images without a background.}
	\label{fig:bad_appels}
\end{figure}

\begin{figure}
	\centering 
	\begin{subfigure}{0.32\linewidth}
		\includegraphics[width=\linewidth, trim={0cm 0cm 0cm 0cm},clip]{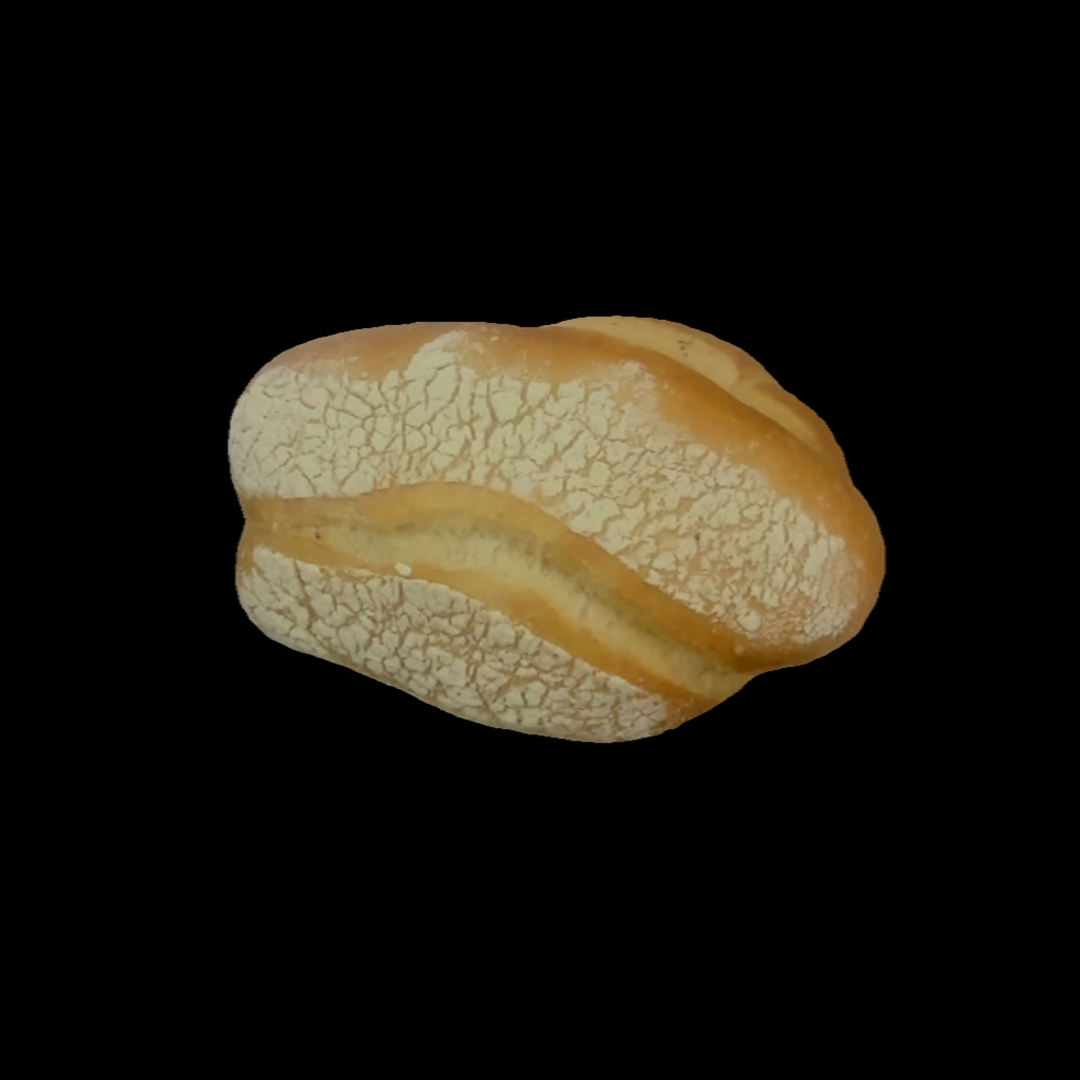}
		\label{fig:examples_1}
	\end{subfigure}\hfil 
	\begin{subfigure}{0.32\linewidth}
		\includegraphics[width=\linewidth, trim={0cm 0cm 0cm 0cm},clip]{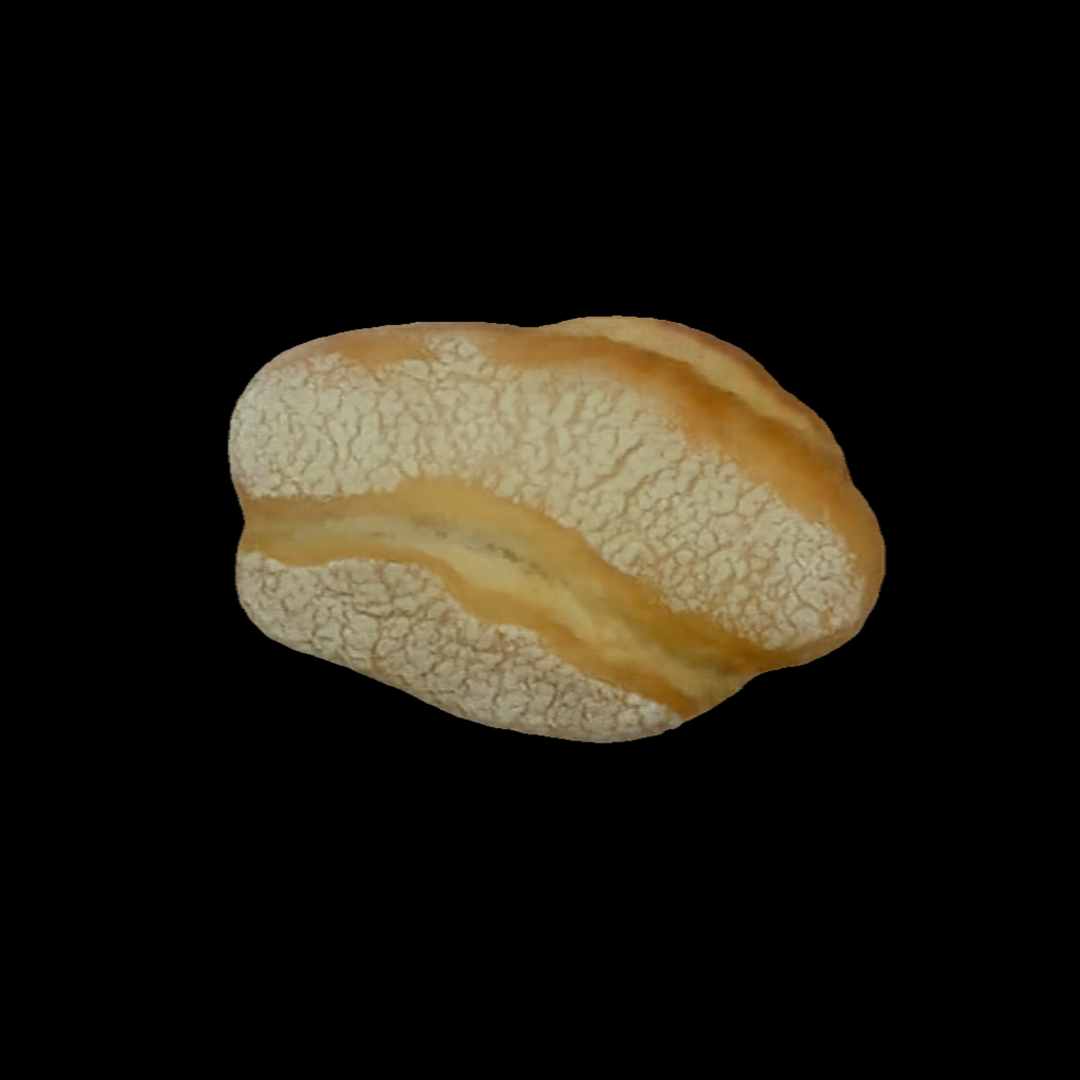}
		\label{fig:examples_2}
	\end{subfigure}\hfil 
	\begin{subfigure}{0.32\linewidth}
		\includegraphics[width=\linewidth, trim={0cm 0cm 0cm 0cm},clip]{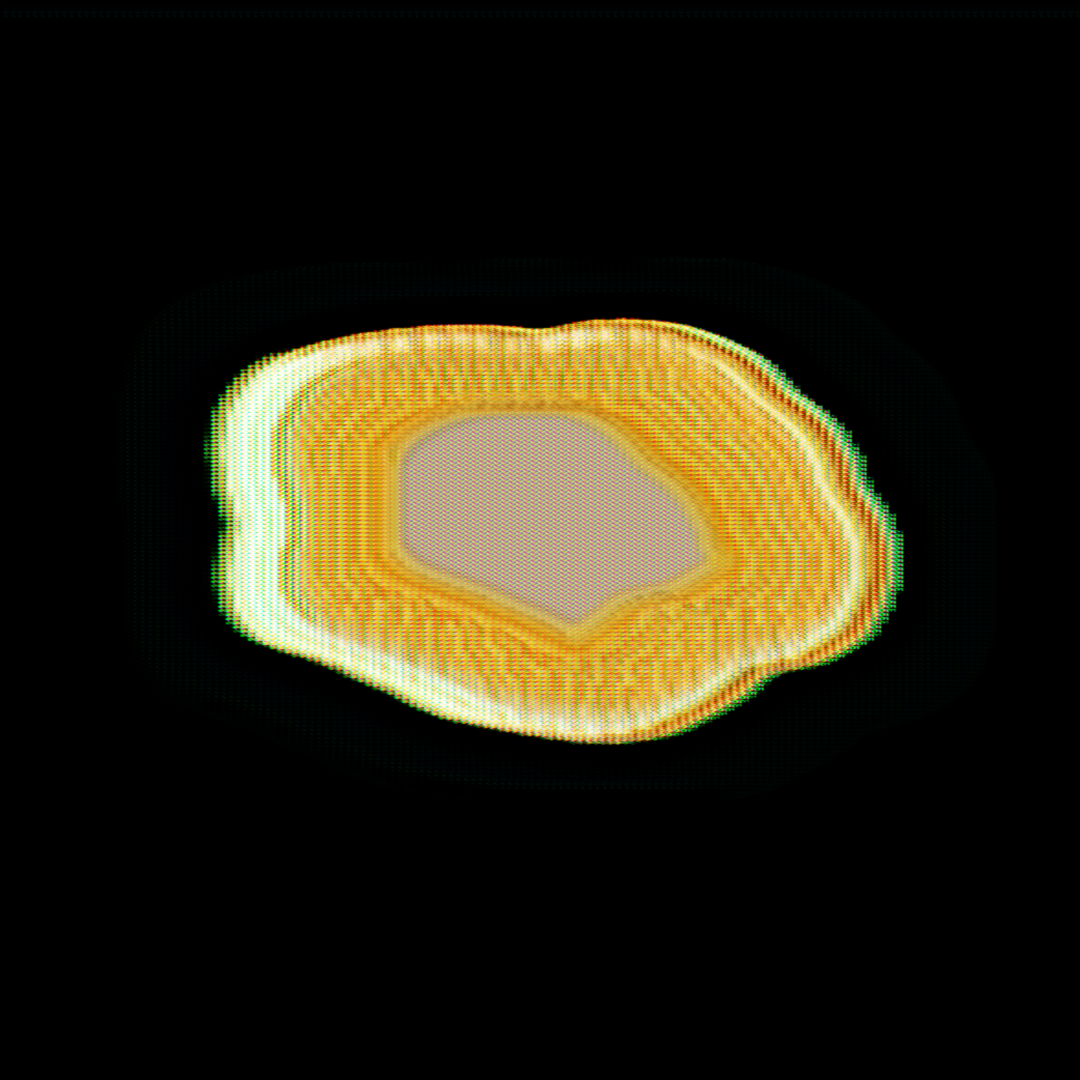}
		\label{fig:examples_3}
	\end{subfigure}
	\caption{Left: Original image of a \emph{Baguettesemmel} (baguette bun). Middle: The corresponding synthetic image generated by our trained pix2pix model. Right: The synthetic image generated by our trained CycleGAN model.}
	\label{fig:bad_baguett}
\end{figure}

To increase the robustness of our detection models against intra-baked good variance, we use generative models to increase the variance of our training data by creating additional training images, from our training set.
We use GAN based generative models pix2pix and CycleGAN to generated additional baked good images.
While their image-to-image translation approach limits the variety of generated images, it allows for greater control over the generation process compared to text-to-image approaches like DreamBooth \cite{DreamBooth}. 
We found this control crucial for preventing training collapse and ensuring the relevancy of generated images,
which we found to be non-trivial when training generative models on our specialized and small training set.
Generative models are trained for \(700\) epochs on our training set at an image scale of \(1024px\).
\newline
\newline
We found that when training our generative models on our original training set, the presence of the drying tray background causes distortions in the generated backed goods.
To alleviate this issue, we remove the background using our segmentation masks.
Although our synthetic images still exhibit minor deconvolution artifacts, this greatly boosts image quality.
Figure \ref{fig:bad_appels} shows the effect removing the drying tray background has on the generated images.
%
%
Moreover, we found significant performance discrepancies between generative models.
Images generated by our trained pix2pix model appear plausible and increase the intra-baked good variance in our training set.
Images generated by our trained CycleGAN model exhibit extreme generation artifacts and are unusable.
When training on our limited dataset, CycleGAN models suffer from catastrophic training collapse, with the discriminator beating the generator early on in training.
This is likely caused by CycleGAN's unpaired image-to-image translation approach, which lends itself well to training on large datasets, but proved detrimental when training on our limited dataset.
Figure \ref{fig:bad_baguett} shows images generated by pix2pix and CycleGAN side by side.
In total, we created \(2042\) supplementary training set images (\(train_{s}\)) from our training set segmentation masks using our trained pix2pix model.
\section{Experiments and Results}
\label{sec:experiments}
\subsection{Experimental Setups}
\label{sec:sec:experimental_setups}
We train object detection models YOLOv9 \cite{YOLOv9} and YOLOv8 \cite{YOLOv8} on our detection task.
Since our application does not need to process videos or work in real-time, we train the largest available model scales, \(YOLOv9e\) and \(YOLOv8x\), with \(58.1\) and \(68.2\) million trainable parameters respectively.
To improve model performance on our specialized task with our limited training data, we train \(YOLOv9e\) and \(YOLOv8x\) pre-trained on the Microsoft COCO dataset \cite{Coco}.
Models are trained with their respective default training hyperparameters and augmentation pipelines, with the addition of online image augmentation pipeline \(DP_{0,04}\).
Models are trained on images standardized such that the longest side is \(1280px\).
To guarantee model convergence, models are trained for \(150\) epochs. 
\cite{Semmeldetector} reported that training models standardized such that the longest side is \(1280px\) maximizes performance.
They also reported that models trained on grayscale images outperform models trained on colored images.
However, in our preliminary tests, we found that training on grayscale images hinders model performance, which is likely due to the increased variety of baked goods in our dataset. 
Therefore, we opt to train our models on colored images.
\subsection{Experiments}
\label{sec:sec:experiments}
We evaluate the effectiveness of \cite{Semmeldetector}'s training approach on our dataset by training our object detection models on training sets \(train_{a}\), \(train_{b}\) and \(2000\) synthetic images created from \(train_{b}\) using our Copy-Paste augmentation pipeline.
The resulting model performances are shown in Table \ref{tab:experiments} as Experiment: \emph{baseline}.
Our model performances are on par with those reported by \cite{Semmeldetector}.
However, our test set images are significantly more challenging for our models and more representative to the true application use case.
The YOLOv9e model outperforms the YOLOv8x model in our \emph{baseline} experiment.
\begin{table}
    \begin{center}
        \small
        \begin{tabular}{lrrrr}
            \toprule
            Experiment & \# Images & Model & mAP@0.5 & max f1-score \\
            \midrule
            baseline & 4520 & YOLOv8x & 0.877 & 0.80@0.747\\
                     &      & YOLOv9e & 0.901 & 0.79@0.727\\
            \midrule
            type-balance & 4995 & YOLOv8x & 0.863 & 0.77@0.658\\
                         &      & YOLOv9e & 0.905 & 0.79@0.646\\
            \midrule
            unknown & 5268 & YOLOv8x & 0.867 & 0.76@0.427\\
                    &      & YOLOv9e & 0.878 & 0.77@0.547\\
            \midrule
            pix2pix & 7780 & YOLOv8x & 0.866 & 0.78@0.694\\
                    &      & YOLOv9e & 0.856 & 0.79@0.587\\
            \midrule
            all-data & 9780 & \textbf{YOLOv8x} & \textbf{0.903} & \textbf{0.82@0.617}\\
                     &      & YOLOv9e & 0.898 & 0.79@0.784\\
            \bottomrule
        \end{tabular}        
        \caption{Experimental results}
        \label{tab:experiments}
    \end{center}
\end{table}
To test whether mitigating the type imbalance or including baked goods classified as \grqq{}unknown\grqq{} improves model performance, we train our models on training sets \(train_{a}\), \(train_{b}\), \(train_{c}\) and \(2000\) synthetic images derived from (\(train_{b}\), \(train_{c}\)); and training sets \(train_{a}\), \(train_{b}\), \(train_{c}\), \(train_{d}\), \(2000\) synthetic images derived from (\(train_{b}\), \(train_{c}\), \(train_{d}\)) respectively.
The resulting performances are shown in Table \ref{tab:experiments} as Experiments: \emph{type-balance} and \emph{unknown}.
Mitigating the type imbalance using oversampling or including baked goods classified as \grqq{}unknown\grqq{} didn't significantly impact model performances for either YOLOv9e or YOLOv8x.
Despite this, we opt to include \grqq{}unknown\grqq{} baked goods in our further experiments to maximize training data variance.
\newline
\newline
To test whether the generative model pix2pix is suitable to increase the variance of small datasets like ours, we train our models on training sets \(train_{a}\), \(train_{b}\), \(train_{c}\) \(train_{s}\) and \(2000\) synthetic images derived from \(train_{s}\).
Since images in \(train_{s}\) posses no relevant background, we use a minor subset of \(train_{b}\) to form image backgrounds during image synthesis.
The resulting performances are shown in Table \ref{tab:experiments} as Experiment: \emph{pix2pix}.
Our experiment demonstrates that, despite minor performance drops, we successfully trained both the YOLOv9e and YOLOv8x models using primarily images generated using pix2pix and our Copy-Paste augmentation pipeline. 
This indicates that pix2pix is capable of retaining all relevant information about our baked goods.
%
%
Finally, to test whether combining all available training data, real or synthetic, benefits our models, we train our models on all available training sets images.
The resulting performances are shown in Table \ref{tab:experiments} as Experiment: \emph{all-data}.
The YOLOv9e model suffers an insignificant performance drop when trained on all training sets images.
This indicates that while pix2pix is capable of reproducing baked goods' visual qualities, its limited generation approach hinders it from significantly increasing training data variance.
The YOLOv8x model shows a significant performance increase, making it outperforming all remaining models, due to its higher maximal F1-score.
That our best-performing model is a YOLOv8 model, and the generally minimal performance difference between our trained YOLOv8x and YOLOv9e models, 
indicates that YOLOv9's increased information retention properties, particularly noticeable at the beginning of the training process, are less significant when fine-tuning large pretrained models.
\section{Conclusions}
\label{sec:conclusions}
In this study, we expanded the work of \cite{Semmeldetector}, by expanding the dataset, training the newly released state-of-the-art YOLOv9 on the baked goods detection task, and tested generative models to create training set images.
We expanded the dataset scope from bread buns to various types of breads and pastries, resulting in a dataset comprising \(2432\) images featuring \(25\) different types of baked goods.
We used the Segment Anything Model (SAM) to semi-automatically annotate our training set images, to facilitate model scalability.
We introduced minor improvements to the Copy-Paste Augmentation pipeline introduced by \cite{Semmeldetector}.
We tested generative models pix2pix and CycleGAN to enrich our small dataset.
We showed that generative models trained on images with repetitive backgrounds can exhibit generation artifacts, and presented a method to mitigate them.
We found that the CycleGAN model was unsuitable for enriching our small dataset, due to its unpaired image-to-image translation approach.
While pix2pix was able to reproduce the visual qualities of our baked goods, it proved to increase training data variance insignificantly.
Our overall best performing model, achieved an \(AP_{0.5}\) of \(90.3\%\) on our test set, which is on par with the results reported by \cite{Semmeldetector} despite our significantly more challenging test set images.
\section{Future Work}
\label{sec:future_work}
Our study could benefit from further research in the following areas:
(1) While \cite{Semmeldetector}'s augmentation pipeline proved invaluable for model training, controlling the relative scale of objects and positioning is still underdeveloped.
(2) Expanding our training data to more images would both bolster model performance and enable us to test more intricate generative models.
(3) We tested the image-to-image generative models pix2pix and CycleGAN to enhance model robustness; however, exploring more intricate generative approaches such as text-to-image could allow us to meaningfully increase training data variance and enhance model robustness.
\section*{Acknowledgments}
We would like to thank Backhaus Müller, local Franconian bakery, for their cooperation and insight.
%
%
\printbibliography
\end{document}